\documentclass[letterpaper]{article} % DO NOT CHANGE THIS
\usepackage{aaai24}  % DO NOT CHANGE THIS
\usepackage{times}  % DO NOT CHANGE THIS
\usepackage{helvet}  % DO NOT CHANGE THIS
\usepackage{courier}  % DO NOT CHANGE THIS
\usepackage[hyphens]{url}  % DO NOT CHANGE THIS
\usepackage{graphicx} % DO NOT CHANGE THIS
\usepackage{natbib}  % DO NOT CHANGE THIS AND DO NOT ADD ANY OPTIONS TO IT
\usepackage{caption} % DO NOT CHANGE THIS AND DO NOT ADD ANY OPTIONS TO IT
\usepackage{algorithm}
\usepackage{algorithmic}

\usepackage{amsmath}
\usepackage{amsfonts}
\usepackage{booktabs}
\usepackage{multirow}

\usepackage{newfloat}
\usepackage{listings}
\DeclareCaptionStyle{ruled}{labelfont=normalfont,labelsep=colon,strut=off} % DO NOT CHANGE THIS
\floatstyle{ruled}
\newfloat{listing}{tb}{lst}{}
\floatname{listing}{Listing}
\title{PA2D-MORL: Pareto Ascent Directional Decomposition based Multi-Objective Reinforcement Learning}
\author {
    Tianmeng Hu\textsuperscript{\rm 1},
    Biao Luo\textsuperscript{\rm 1} \thanks{Corresponding author.}
}
\affiliations {
    \textsuperscript{\rm 1}Central South University\\
    tianmeng0824@163.com, biao.luo@hotmail.com
}

\usepackage{bibentry}
\begin{document}

\maketitle

\begin{abstract}
    Multi-objective reinforcement learning (MORL) provides an effective solution for decision-making problems involving conflicting objectives. However, achieving high-quality approximations to the Pareto policy set remains challenging, especially in complex tasks with continuous or high-dimensional state-action space. In this paper, we propose the Pareto Ascent Directional Decomposition based Multi-Objective Reinforcement Learning (PA2D-MORL) method, which constructs an efficient scheme for multi-objective problem decomposition and policy improvement, leading to a superior approximation of Pareto policy set. The proposed method leverages Pareto ascent direction to select the scalarization weights and computes the multi-objective policy gradient, which determines the policy optimization direction and ensures joint improvement on all objectives. Meanwhile, multiple policies are selectively optimized under an evolutionary framework to approximate the Pareto frontier from different directions. Additionally, a Pareto adaptive fine-tuning approach is applied to enhance the density and spread of the Pareto frontier approximation. Experiments on various multi-objective robot control tasks show that the proposed method clearly outperforms the current state-of-the-art algorithm in terms of both quality and stability of the outcomes. 
  
  \end{abstract}
  
  \section{Introduction}
  
  Deep Reinforcement Learning (RL) is a promising approach for solving decision-making and planning problems, and it has found widespread applications in various fields such as robot control \cite{schulman2017proximal}, autonomous driving \cite{feng2023dense}, gaming \cite{silver2017mastering}, and recommender systems \cite{do2022online}. In Deep RL, an agent's behavior is guided by a reward function that defines the optimization objective. However, decision-making problems can involve multiple conflicting objectives, which is an important challenge for real-world applications of RL \cite{dulac2021challenges}. For instance, an autonomous driving task may have two objectives: vehicle speed and passenger comfort, while a robot control task may involve forward speed and energy efficiency. In such multi-objective decision-making problems, a single optimal policy does not exist because conflicting objectives cannot be simultaneously optimized. Instead, there exists a set of Pareto-optimal policies. Adapting to requirements and application scenarios, human users must select a trade-off among multiple objectives that reflects their objective preferences, and different preferences align with different Pareto-optimal policies.
  
  In this context, multi-objective reinforcement learning has attracted increasing attention and many advanced methods have been proposed. Among them, the multi-policy MORL \cite{chen2019meta,yang2019generalized, xu2020prediction, hayes2022practical} is dedicated to searching a non-dominated policy set consisting of a series of policies aligned to different objective preferences. Compared with the single-policy approach \cite{chen2021emorl, ijcai2022p476, kyriakis2022pareto}, the multi-policy approach has great advantages in terms of applicability and flexibility, despite being more complicated. On the one hand, by visualizing the non-dominated policy set, users are free to select satisfactory policies to suit various demands. On the other hand, the multi-policy MORL can also be combined with methods that focus on real-world applications, such as safe RL \cite{horie2019multi}, since the policies that satisfy the safety constraints can be easily identified from the non-dominated policy set. 
  
  Recently, deep policy gradient methods and the actor-critic architecture have also been introduced into multi-objective reinforcement learning to address complex tasks with continuous or high-dimensional state-action spaces. The Pareto policy adaptation (PPA) \cite{kyriakis2022pareto} method designs a loss function based on a multi-objective gradient and attempts to search for a single solution in a given preference distribution. It is easy to implement, but the policy needs to be fine-tuned or retrained when preferences change. In contrast, the Prediction-Guided MORL (PGMORL) \cite{xu2020prediction} method employs an intuitive prediction model and an evolutionary architecture to address multi-objective robot control problems and achieve high-quality sets of non-dominated policies. However, the limitations of the prediction model can negatively affect the performance of policies and the stability of the results. Therefore, instead of introducing predictive models, we explore completely different solutions.
  
  In this paper, we focus on multi-policy MORL and propose a novel policy gradient-based approach to obtain high-quality approximations of the Pareto policy set, outperforming current state-of-the-art approaches. The proposed method leverages the Pareto ascent direction to decompose a multi-objective decision-making problem into a series of single-objective decision-making problems, and achieves parallel improvement of multiple policies with different directions in an evolutionary framework. A non-dominated policy set is maintained during training and gradually approximates the Pareto frontier. The main contributions of this paper can be summarized as follows:
  
  \begin{itemize}
  \item We propose a novel multi-policy MORL method for Pareto policy set approximation via Pareto Ascent Directional Decomposition (PA2D-MORL). For a non-Pareto optimal policy, the method finds a decomposition of the original multi-objective problem by solving the Pareto ascent direction that improves all objectives simultaneously and avoids introducing any prior objective preference. In this way, an optimization direction that depends on the policy parameters can be determined automatically, avoiding the use of human-designed prediction models.
  \item We propose a partitioned greedy randomized approach to select policies for updating under an evolutionary framework, which helps the policies to move towards a higher-performance and wider objective space. This encourages effective exploration and exploitation, and avoids getting stuck in the same local optimum for long term. 
  \item We propose a Pareto adaptive fine-tuning method, which selects several policies to be fine-tuned according to the current distribution of non-dominated policies in the objective space, thus improving the quality of the Pareto set approximation.
  \item We evaluate the proposed method in seven MuJoCo environments that are modified for multi-objective tasks, and achieve state-of-the-art performance in terms of both the quality of the Pareto set approximation and the stability of the results. 
  \end{itemize}

  \section{Related Work}

  As reinforcement learning is being used to solve complex real-world decision-making problems, many of which have multiple conflicting objectives, MORL is an effective way to solve such problems. Current MORL methods can be broadly divided into two categories: single-policy and multi-policy methods. Some relevant work is presented below.

  \subsection{Single-Policy Methods}
  Single-policy methods transform multi-objective problems into single-objective problems. A typical approach is to use specific weights to scalarize the multi-objective rewards into a scalar reward, and then run the reinforcement learning algorithm to find an optimized policy. For example, Duan et al. artificially designed a synthetic multi-objective reward function and applied policy gradient DRL methods to several continuous control problems \cite{duan2016benchmarking}. Chen et al. applied MORL to hyper-parameter optimization problems, designing a scalarization function combining accuracy and latency as a synthetic reward to guide policy updating \cite{chen2021emorl}. 
  Furthermore, the lexicographic-MORL \cite{ijcai2022p476} optimizes the rewards lexicographically according to the importance and constraints of different objectives. The Pareto policy adaptation \cite{kyriakis2022pareto} method uses a gradient solver to calculate the policy optimization direction and modify the loss function. This method can learn a single optimized policy for a given preference distribution.
  
  Typically, single-policy MORL employ some prior information to find a single solution that meets the requirements. However, these methods have several problems: 1) requirements and application scenarios may change, making the policy need to be fine-tuned or re-trained; 2) user preferences may be abstract and difficult to quantify accurately. An effective way to solve above problems is to find a coverage set of optimized policies, allowing users to select policies based on their preferences and demands. 
  
  \subsection{Multi-Policy Methods}
  Multi-policy MORL learns a set of policies, which contain optimized policies corresponding to different user preferences, and presents them to the user for selection. An intuitive scheme is the outer-loop methods, like the scalarized Q-learning \cite{mossalam2016multi}. These methods use an outer loop to iterate over the parameters of the scalarization function and apply a single-objective RL for each parameter. Other approaches aim to avoid explicit iterative search, for example, Pareto-Q-Learning \cite{van2014multi} modifies the original Q-learning method to store multiple Pareto optimal values for each state-action pair. To deal with the problem of high-dimensional state spaces, some methods have extended Deep Q-Networks (DQN) \cite{mnih2015human} to multi-objective domains, such as Pareto-DQN \cite{reymond2019pareto} and the multi-objective Q-network \cite{abels2019dynamic}. The latter has been improved by Envelope Q-Learning \cite{yang2019generalized}, which performs envelope updates using the convex envelope of the policy frontier, allowing the algorithm to converge faster. MO-MIX \cite{hu2023mo} extends multi-agent value decomposition to solve multi-objective multi-agent cooperative decision-making problems. 
  Further, deep policy gradient methods have been introduced to deal with continuous action spaces. 
  Meta-MORL \cite{chen2019meta} first trains a meta-policy as the best initial policy, then performs fine-tuning on the meta-policy to obtain optimized policies corresponding to different preferences. PGMORL \cite{xu2020prediction} constructs an evolutionary framework and selects policies based on a prediction model, these policies are then updated using the Proximal Policy Optimization (PPO) \cite{schulman2017proximal} algorithm. 
  
  PGMORL is the current state-of-the-art multi-policy algorithm. However, the method relies on an intuitive prediction model for predicting potential improvements in policies. The accuracy of this model cannot be guaranteed, which may adversely affect the performance and stability of results. It also suffers from the long-term local minima problem in some tasks. In addition, developing a highly precise prediction model is challenging and can involve significant additional computational cost, which is not an ideal solution.
  
  To address the above problems, our work solves the optimization direction of the policies mathematically instead of using a prediction model, which gives the method a firmer mathematical foundation and improves the stability of the results. The proposed method can guide the policies to a better performance space and generate a higher-quality Pareto policy set approximation.

  \section{Preliminaries}

  \subsection{Multi-objective Decision-making}
  A multi-objective decision-making problem involves optimizing multiple objectives:
  \begin{equation}
    \label{moo}
    \max _{\pi_\theta} \mathbf{F}\big( \pi_\theta \big)= 
    \max _{\pi_\theta}
    \Big[f_{1}\big(\pi_\theta \big), \ldots, f_{m}\big(\pi_\theta \big)\Big],
  \end{equation}
  where $ f_{1}, \ldots, f_{m} $ denotes the objective functions corresponding to the $m$ objectives, and $\pi_\theta $ denotes the parameterized policy.
  
  Further, a multi-objective sequential decision-making problem can be formulated as a Multi-Objective Markov Decision Process (MOMDP) consisting of a tuple $ \mathcal{G}=\langle\mathcal{S}, \mathcal{A}, \mathcal{P}, \mathbf{r}, \gamma\rangle $, where $ \mathcal{S} $ is the state space, $ \mathcal{A} $ is the action space, and the multi-objective reward $ \mathbf{r} $ is the vector consisting of the individual objective rewards.
  An RL agent receives the environment state $ s \in \mathcal{S} $ and selects the action $ a \in \mathcal{A} $ according to the policy $ \pi\left(s\right)$. The state of the environment is transformed according to the transition function $ \mathcal{P}\left(s^{\prime} \mid s, a\right) $.
  
  Reinforcement learning method finds the optimal policy ${\pi}^{\ast}$ by maximizing expected return. The return is a discounted sum of future rewards. For MORL, the multi-objective return can be represented as a vector $\mathbf{G}$, in which the return on objective $i$ is defined as: 
  \begin{equation}
    \label{Gi}
    {G}_{i}=\sum_{t=0}^{\infty} \gamma^{t} {r}_{i}\left(s_{t}, {a}_{t}\right),
  \end{equation}
  where $\gamma$ is a discount factor that trade off long-term returns against short-term rewards, and ${r}_{i}$ denotes the reward corresponding to objective $i$. Accordingly, the expected return is defined as:
  \begin{equation}
    J_{i}^{\pi_\theta}=E_{\boldsymbol{\tau} \sim \pi_{\theta}(\boldsymbol{\tau})}[{G}_{i}(\boldsymbol{\tau})]=\int_{\boldsymbol{\tau} \sim \pi_\theta(\boldsymbol{\tau})} \pi_\theta(\boldsymbol{\tau}) {G}_{i}(\boldsymbol{\tau}) \mathrm{d} \boldsymbol{\tau} ,
  \end{equation} 
  where $\boldsymbol{\tau}$ denotes a trajectory collected by interaction using the policy $\pi_\theta$. For MORL, we use $J_i$ as the objective function of objective $i$.
  
  \subsection{Pareto Optimality}
  Let $\mathbf{J}^{\pi} = \left[J _{1}^{\pi}, \ldots, J _{m}^{\pi} \right]^{\top} $ be the multi-objective expected return of policy $\pi$. 
  The policy $\pi_{1}$ dominates the policy $\pi_{2}$ if and only if $ J_{i}^{\pi_{1}} \geq J_{i}^{\pi_{2}} $ for each $i \in \left( 1, \ldots, m \right)$ and $ \mathbf{J}^{\pi_{1}} \neq \mathbf {J}^{\pi_{2}} $.
  In general, multiple objectives cannot be optimized simultaneously, so there exists a set of optimized policies consisting of policies that are not dominated by any other policy, i.e., the Pareto policy set. Accordingly, there exists a Pareto frontier, which is a mapping of the Pareto policy set into the objective space.

  \begin{figure}[t]
    \centering
    \includegraphics[width=0.94\columnwidth]{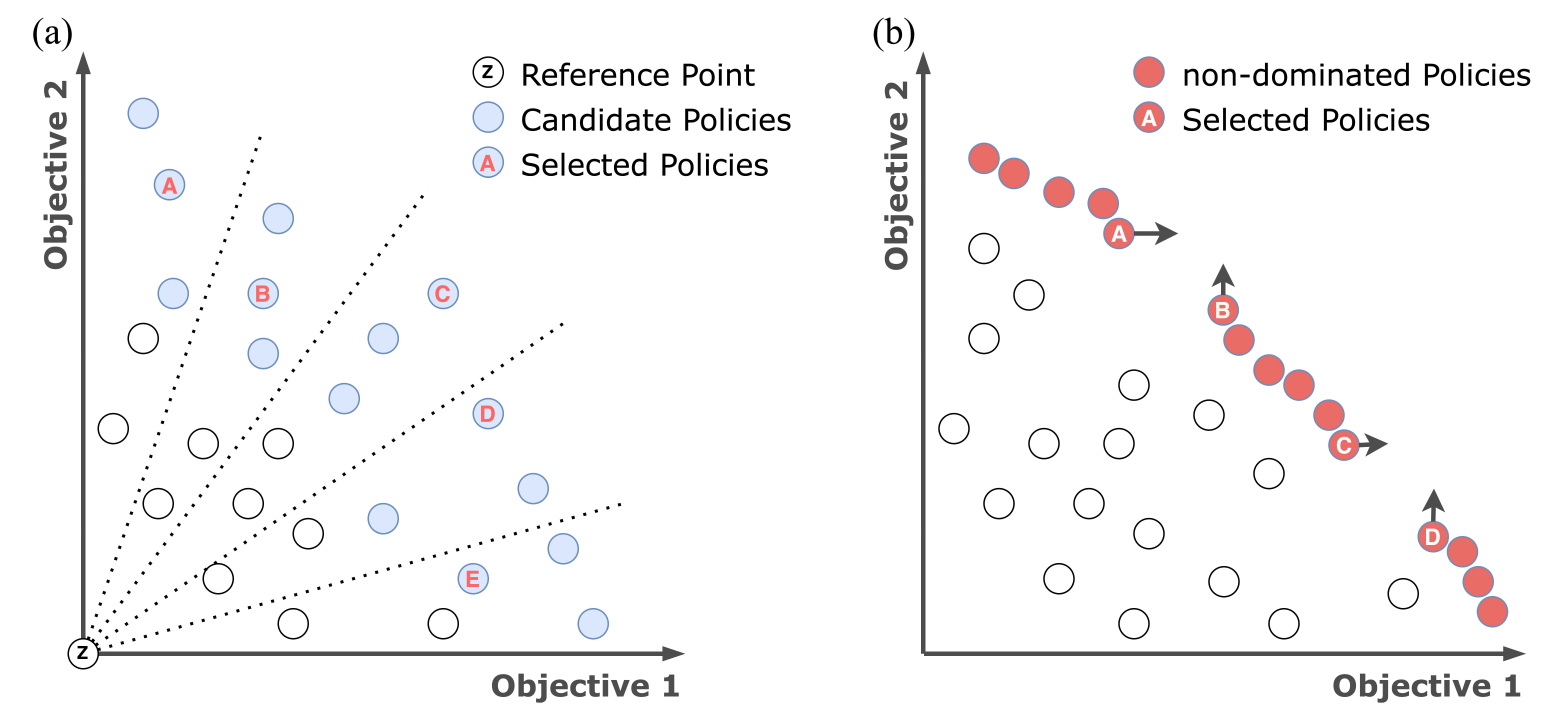} % Reduce the figure size so that it is slightly narrower than the column.
    \caption{Illustration of the PGR and PA-FT methods. (a) A portion of the better performing policies in each partition is selected as candidate policies, and then one is randomly selected from the candidates. (b) The larger missing regions in the current Pareto frontier approximation are identified. Policies around the missing region are selected and fine-tuned. }
    \label{schematic}
  \end{figure}

  \section{Method}
  
  In this section, we propose the Pareto ascent directional decomposition based MORL method (PA2D-MORL) to obtain high quality approximations to the Pareto policy set. The three main aspects of our method are elaborated below, and the pseudo-code of the algorithm is shown in Algorithm~\ref{alg1}. 
  
  \subsection{Pareto Ascent Directional Decomposition}
  
  A Pareto-optimal solution to a multi-objective optimization problem will also be an optimal solution to a single-objective optimization problem whose objective function is an aggregation of all the objective functions of the original problem \cite{zhang2007moead}. This indicates that a multi-objective problem can be decomposed into multiple single-objective problems, and thus the approximation of Pareto policy set can be decomposed into the optimization of these single-objective problems. We employ the weighted sum approach to scalarize the multi-objective returns:
  \begin{equation}
    \label{weighted-sum}
  {J}^{\pi_\theta}(\boldsymbol{\omega})=\boldsymbol{\omega}^{\top} \mathbf{J}^{\pi_\theta}=\sum_{i=1}^m \omega_i J_i^{\pi_\theta} 
  \quad s.t. \quad \sum_{i=1}^m \omega_i = 1.
  \end{equation}
  Intuitively, the weight vector $\boldsymbol{\omega}$ denotes the objective preference. Applying various weights can produce a series of single-objective problems whose optimal solutions are individually aligned to different points on the Pareto frontier.
  
  For a multi-objective problem with $m$ objectives, there are $m$ gradient directions: 
  \begin{equation}
    \label{pg}
  \nabla_\theta J_{i}^{\pi_\theta}=
  \int_{\tau \sim \pi_\theta(\tau)} \nabla_\theta \pi_\theta(\tau) {G}_{i}(\tau) \mathrm{d} \tau.
  \end{equation}
  Then, for a decomposition problem based on weights $\boldsymbol{\omega}$, the aggregated policy gradient is: 
  \begin{equation}
    \label{mopg}
  \nabla_{\theta} {J}^{\pi_\theta}(\boldsymbol{\omega})
   =\sum_{i=1}^m \omega_i \nabla_{\theta} J_i^{\pi_\theta} 
  \quad s.t. \quad \sum_{i=1}^m \omega_i = 1,
  \end{equation}
  which assigns an optimization direction towards the Pareto frontier for the current policy, and allows the implementation of RL. Although it is possible to decompose a multi-objective problem into many single-objective problems by dense sampling in the weight distribution, it is inefficient and impractical to apply separate deep RL algorithms to each of these decomposed problems for optimization. Approximating the Pareto policy set remains challenging, and how to choose the policy and the decomposed problem is critical. 
  
  We argue that there are two aspects to consider: 1) the optimization direction chosen for a non-Pareto policy should efficiently guide the policy to approximate the aligned Pareto policy, which can be achieved by choosing a direction that improves all objectives simultaneously; 2) this optimization direction should avoid introducing prior policy preferences. Therefore, we leverage the Pareto ascent direction, which is a multi-objective common ascent direction with minimum norm.

  \begin{algorithm}[tb]
    \caption{PA2D-MORL}
    \label{alg1}
    \textbf{Input}: total generations: $M$; PA-FT generations: $M_{ft}$; iterations per generation: $m$; warmup iterations: $m_w$; initialized non-dominated policy set $P_n$.
    % \textbf{Parameter}: Optional list of parameters\\
    \begin{algorithmic}[1] %[1] enables line numbers
    \STATE \textbf{Warmup}: 
    \STATE Generate $p$ randomly initialized policies and $p$ evenly distributed weights $\boldsymbol{\omega}$. 
    \STATE Update each policy for $m_w$ iterations according to \eqref{mopg} to generate the first generation of policy population $P$
    \STATE \textbf{Evolution}: 
    \FOR{$generation = 1 \dots M$}
      \IF {$generation < M_{ft}$}
      \STATE $p_a = p, p_b=0 .$
      \ELSE
      \STATE $p_a = p_b = \frac{1}{2}p .$
      \ENDIF
      \STATE Select $p_a$ policies from $P$ by the PGR approach. 
      \FOR{$k = 1 \dots p_a$}
        \STATE Calculate the policy gradient of $\pi_k$ according to \eqref{pg} and solve \eqref{min-norm} to obtain $\ \boldsymbol{\alpha^{\ast}}$.
        \STATE Update $\pi_k$ for $m$ iterations by \eqref{mopg}, where $\boldsymbol{\omega} = \boldsymbol{\alpha^{\ast}}$.
      \ENDFOR
      \IF {$generation \geq M_{ft}$}
      \STATE Select and update $p_b$ policies from $P_n$ by the PA-FT approach.
      \ENDIF
      \STATE Update $P_n$ and $P$ with new policies.
    \ENDFOR
    \end{algorithmic}
    \textbf{Output}: Pareto policy set approximation $P_n$.
  \end{algorithm}

  \subsubsection{Pareto Ascent Direction}
  
  We can derive the Pareto ascent direction \cite{desideri2012multiple} by considering the following optimization problem: 
  \begin{equation}
    \label{min-norm}
  \min _{\alpha_1, \ldots, \alpha_m}\left\{\left\|\sum_{i=1}^m \alpha_i \nabla_{{\theta}} J_i^{\pi_\theta} \right\|_2^2
   s.t.  \sum_{i=1}^m \alpha_i=1 , \alpha_i \geq 0 (\forall i) \right\}.
  \end{equation}
  There are two possible situations:
  \begin{itemize}
    \item If the optimal solution $\boldsymbol{\alpha}^{\ast}$ of the above problem leads to $\sum_{i=1}^m \alpha_i^{\ast} \nabla_{\theta} J_i^{\pi_\theta} = 0$, then $\pi_\theta $ is a Pareto stationary policy. It is the first-order necessary condition for Pareto optimality. 
    \item Otherwise, the Pareto-stationarity condition does not hold, $\pi_\theta $ is not a Pareto-optimal policy, and $\sum_{i=1}^m \alpha_i^{\ast} \nabla_{\theta} J_i^{\pi_\theta}$ indicates a common ascent direction on all objectives, which we call the Pareto ascent direction.
  \end{itemize}
  
  Theoretically, all objectives are improved by the same amount on the Pareto ascent direction since the gradient vectors for different objectives have equal projections on it \cite{desideri2012multiple}. We therefore employ the Pareto ascent direction to guide the optimization of the non-Pareto policies, which means applying $\boldsymbol{\omega} = \boldsymbol{\alpha^{\ast}}$ and calculating the scalarized return based on \eqref{weighted-sum}. This return can be optimized by policy gradient approaches such as the PPO algorithm. 
  
  For the optimization problem \eqref{min-norm}, a simple stochastic gradient descent can be applied to solve it. We can use a projection function \cite{duchi2008efficient} to handle the constraints: 
  \begin{equation}
    \boldsymbol{\alpha}^{t+1}=\mathcal{P}_{\mathcal{C}}\left(\boldsymbol{\alpha}^t - \eta_t \nabla_{\boldsymbol{\alpha}} f\left(\boldsymbol{\alpha}^t\right)\right),
  \end{equation}
  where $ f\left(\boldsymbol{\alpha}\right) = \left\|\sum_{i=1}^m \alpha_i \nabla_{{\theta}} J_i^{\pi_\theta} \right\|_2^2 $, and $\eta$ is the step size. $\mathcal{P}_{\mathcal{C}}\left(\boldsymbol{\alpha}\right)$ is the projection function that is used to find the projection of $\boldsymbol{\alpha}$ onto the probabilistic simplex to ensure that the constraints $\sum_{i=1}^m \alpha_i=1, \alpha_i \geq 0 (\forall i)$ are satisfied. 
  
  In particular, for the case involving two objectives, i.e., $\alpha_1 + \alpha_2 = 1$, an analytical solution can be found by geometric analysis. The minimum norm vector is either perpendicular to the difference of the two gradient vectors or coincides with one of the gradients, therefore: 
  \begin{equation}
  \alpha_{1}^{\ast}=\max \left(
  \min \left(
  \frac{\left(\nabla_{\theta} J_2^{\pi_\theta} - \nabla_{\theta} J_1^{\pi_\theta} \right)^{\top} \nabla_{\theta} J_2^{\pi_\theta} }
  {\left\| \nabla_{\theta} J_1^{\pi_\theta}    -  \nabla_{\theta} J_2^{\pi_\theta} \right\|_2^2}, 1
  \right), 0
  \right).
  \end{equation}
  % Then, the optimal solution of \eqref{min-norm} is $ \hat{\boldsymbol{\alpha}} = \left[ \hat{\alpha}_1, 1 - \hat{\alpha}_1 \right]^{\top} $.

  % \begin{figure}[t]
  %   \centering
  %   \includegraphics[width=0.94\columnwidth]{schematic} % Reduce the figure size so that it is slightly narrower than the column.
  %   \caption{Illustration of the PGR and PA-FT methods. (a) A portion of the better performing policies in each partition is selected as candidate policies, and then one is randomly selected from the candidates. (b) The larger missing regions in the current Pareto frontier approximation are identified. Policies around the missing region are selected and fine-tuned. }
  %   \label{schematic}
  % \end{figure}

  \subsection{Partitioned Greedy Randomized Policy Selection}
  
  We divide the training process into multiple generations and maintain a population of policies. In each generation, a series of policies are selected and updated in parallel. Further, we propose a partitioned greedy randomized (PGR) approach to select the policies that need to be updated. Specifically, we divide the objective space into $n$ regions based on angular range and use a distance metric to rank the policies that fall in each of these regions. The distance metric is defined as the distance of a policy from the reference point in the objective space:
  \begin{equation}
  \mathcal{D} = \| \mathbf{J}^{\pi} - \mathbf{Z} \|_2,
  \end{equation}
  where $\mathbf{Z}$ is the reference point which is dominated by all possible policies. Then, the best $k$ policies in each region are identified and one of them is randomly selected. 
  Next, the $n$ selected policies are updated according to their respective Pareto ascent directions. The intermediate policies from the optimization process are stored and are used to update the set of non-dominated policies.
  
  PGR encourages exploration and exploitation and strikes a balance between them. On the one hand, by greedily selecting policies from different partitions, the policy population can be moved towards a wider and higher-performance objective space. On the other hand, it is important to introduce randomness, which helps to jump out of the local optimum and avoid the long-term local minima problem.

  \subsection{Pareto Adaptive Fine-tuning}
  
  Multi-generational optimisation following the Pareto ascent direction can lead to high-performance policies corresponding to different preferences, but does not guarantee a uniform approximation of the Pareto frontier. To address this problem, we propose the Pareto adaptive fine-tuning (PA-FT) method, which is embedded into the training process. 
  
  Let $\mathbb{N}$ denote the non-dominated policy set, and the approximation of the Pareto front $\mathbb{J}=\left \{ \mathbf{J}^{\pi} \mid {\pi}\in \mathbb{N} \right \} $ denote the mapping of the non-dominated policies in the objective space, which is a series of discrete points in the m-dimensional space. The distribution of these points is non-uniform, hence we focus on large missing spaces and try to cover them. 
  To find these spaces, a nearest neighbor search is applied to identify the nearest neighbors of each point in $\mathbb{J} $, as well as the nearest distances.
  The policies corresponding to the $n$ pairs of points with the largest nearest distance will be selected. Two policies in each pair will be assigned opposite optimization directions to cover the missing spaces in the Pareto approximation frontier. In addition, the optimal policy on the $i$-th objective is also selected and updated according to the direction of $\nabla_{{\theta}} J_i^{\pi}$ so that the Pareto approximation frontier extends towards the ends.
  
  During training, Pareto ascent directional optimization and PA-FT are used in combination. Some policies are selected using the PGR method and optimized on the Pareto ascent direction, which pushes the non-dominated policies to a better performance space. Some other policies are selected and updated by PA-FT method, resulting in a denser approximation of the Pareto frontier.

  \section{Experiments}

  % \begin{figure}[t]
  %   \centering
  %   \includegraphics[width=0.97\columnwidth]{hv_sp} 
  %   \caption{Hypervolume and sparsity curves on the Walker environment. The light-colored parts show the standard deviation. Data are based on 6 independent runs.}
  %   \label{hv_sp}
  % \end{figure}

  \subsection{Evaluation Metrics}
  
  Two metrics are used in the experiments to evaluate the quality of the results: the hypervolume metric (HV) and the sparsity metric (SP). Briefly, the HV is a comprehensive metric reflecting the convergence, spread, and homogeneity of the approximation set, while the SP metric reflects the density of the Pareto frontier approximation. 
  The hypervolume is the volume of the space enclosed by nondominated policies $\mathbb{N}$ and the reference point in the objective space:
  \begin{equation}
    \operatorname{HV} =\delta\left(\bigcup_{i=1}^{\mid \mathbb{N} \mid} \mathcal{V}_{i}\right),
  \end{equation}
  where $\delta $ denotes the Lebesgue measure, which measures the $n$-dimensional volume in $n$-dimensional Euclidean space. Specifically, in 2-dimensional space, $\delta $ is a measure of area. $\mathcal{V}_{i}$ denotes the space enclosed by the $i$-th policy in $\mathbb{N}$ with the reference point in the objective space. HV is a Pareto-compliant metric \cite{Cardona2022evo} that if an approximation set $\mathbb{N}_1$ dominates another set $\mathbb{N}_2$, then $ \operatorname{HV} (\mathbb{N}_1) > \operatorname{HV} (\mathbb{N}_2) $. Therefore, we use HV to evaluate the comprehensive quality of the Pareto policy set approximation, and a better approximation set will have a higher HV. 
  The sparsity metric is defined by the distance between discrete points in the Pareto frontier approximation $\mathbb{J}$. Let $\mathbb{J}_i$ be a list of all values on the ith objective in $\mathbb{J}$ and $\widetilde{\mathbb{J}_i}$ be the ordered list. Then, 
  \begin{equation}
    \operatorname{SP} =\frac{1}{|\mathbb{J}|-1} \sum_{i=1}^{m} \sum_{j=1}^{|\mathbb{J}|-1}
  \left( \widetilde{\mathbb{J}_i}(j+1) - \widetilde{\mathbb{J}_i}(j) \right )^{2}.
  \end{equation}
  The smaller the SP metric, the denser the approximation set.

  \begin{table*}[t]
    \centering
    
    \begin{tabular}{c c c c c c c}
      \toprule
        & & PA2D-MORL & PA2D-ablated & PGMORL & PFA & MOEA/D\\
      \midrule
        \multirow{2}{*}{Walker2d} & HV $\uparrow(\times 10^6)$ & $\mathbf{5.743\pm0.121}$ & $5.320\pm0.186$ & $4.849\pm0.558$ & $4.329\pm0.553$ & $4.612\pm0.545$\\
                                  & SP $\downarrow(\times 10^4)$& $\mathbf{0.014\pm0.006}$ & $0.180\pm0.096$ & $0.021\pm0.018$ & $0.309\pm0.225$ & $0.710\pm0.285$\\
      \midrule
        \multirow{2}{*}{Humanoid} & HV $\uparrow(\times 10^6)$& $\mathbf{51.23\pm2.66}$ & $42.93\pm4.14$ & $44.75\pm5.81$ & $40.55\pm5.02$ & $46.35\pm7.33$\\
                                  & SP $\downarrow(\times 10^4)$& $\mathbf{0.133\pm0.031}$ & $0.274\pm0.177$ & $0.255\pm0.121$ & $0.715\pm0.516$ & $2.871\pm1.342$\\  
      \midrule                                   
        \multirow{2}{*}{HalfCheetah} & HV $\uparrow(\times 10^6)$& $\mathbf{5.787\pm0.020}$ & $5.741\pm0.053$ & $5.782\pm0.018$ & $5.765\pm0.081$ & $5.739\pm0.075$\\
                                  & SP $\downarrow(\times 10^4)$& $0.026\pm0.013$ & $0.106\pm0.035$ & $\mathbf{0.022\pm0.015}$ & $0.548\pm0.209$ & $0.679\pm0.295$\\
      \midrule
        \multirow{2}{*}{Hopper-2} & HV $\uparrow(\times 10^6)$& $\mathbf{22.09\pm0.57}$ & $21.30\pm0.68$ & $19.10\pm2.41$ & $20.61\pm4.31$ & $20.73\pm1.17$\\
                                  & SP $\downarrow(\times 10^4)$& $\mathbf{0.503\pm0.107}$ & $1.868\pm0.389$ & $0.559\pm0.529$ & $4.485\pm2.219$ & $2.346\pm0.672$\\
      \midrule
        \multirow{2}{*}{Ant} & HV $\uparrow(\times 10^6)$& $\mathbf{6.814\pm0.167}$ & $6.242\pm0.294$ & $6.283\pm0.277$ & $6.209\pm0.464$ & $6.233\pm0.477$\\
                                  & SP $\downarrow(\times 10^4)$& $\mathbf{0.209\pm0.019}$ & $0.351\pm0.047$ & $0.832\pm0.457$ & $1.021\pm0.554$ & $1.696\pm0.581$\\
      \midrule
        \multirow{2}{*}{Swimmer} & HV $\uparrow(\times 10^4)$& $\mathbf{3.187\pm0.056}$ & $2.965\pm0.336$ & $2.566\pm0.595$ & $2.392\pm0.467$ & $2.323\pm0.531$\\
                                  & SP $\downarrow(\times 10^1)$& $\mathbf{0.550\pm0.207}$ & $0.603\pm0.241$ & $0.917\pm0.862$ & $1.976\pm0.582$ & $2.601\pm1.094$\\
      \midrule
        \multirow{2}{*}{Hopper-3} & HV $\uparrow(\times 10^{10})$& $\mathbf{3.889\pm0.191}$ & $3.759\pm0.277$ & $3.766\pm0.254$ & $-$ & $3.681\pm0.434$\\
                                  & SP $\downarrow(\times 10^3)$& $\mathbf{0.021\pm0.013}$ & $0.106\pm0.052$ & $0.032\pm0.011$ & $-$ & $0.642\pm0.215$\\

      \bottomrule
    \end{tabular}
  
  \caption{Evaluation results for all algorithms. The experiments are based on 7 MuJoCo environments. Both the average and standard deviation of the HV and SP metrics are reported. The modified PFA is difficult to apply to the 3-objective task and is not tested on Hopper-3. All data are based on 6 independent runs.}
  \label{allresults}
  \end{table*}

  \begin{figure}[t]
    \centering
    \includegraphics[width=0.97\columnwidth]{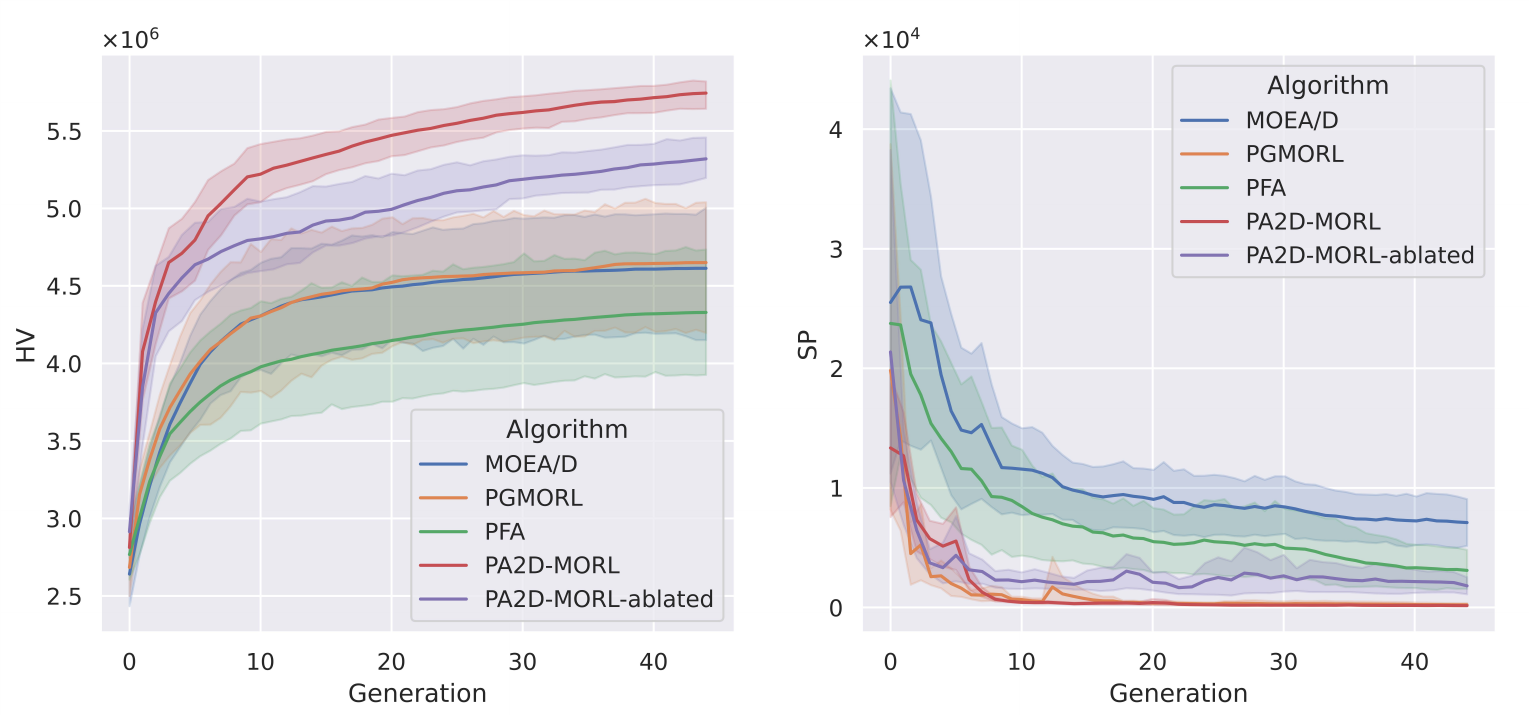} 
    \caption{Hypervolume and sparsity curves on the Walker environment. The light-colored parts show the standard deviation. Data are based on 6 independent runs.}
    \label{hv_sp}
  \end{figure}

  \begin{figure}[t]
    \centering
    \includegraphics[width=0.97\columnwidth]{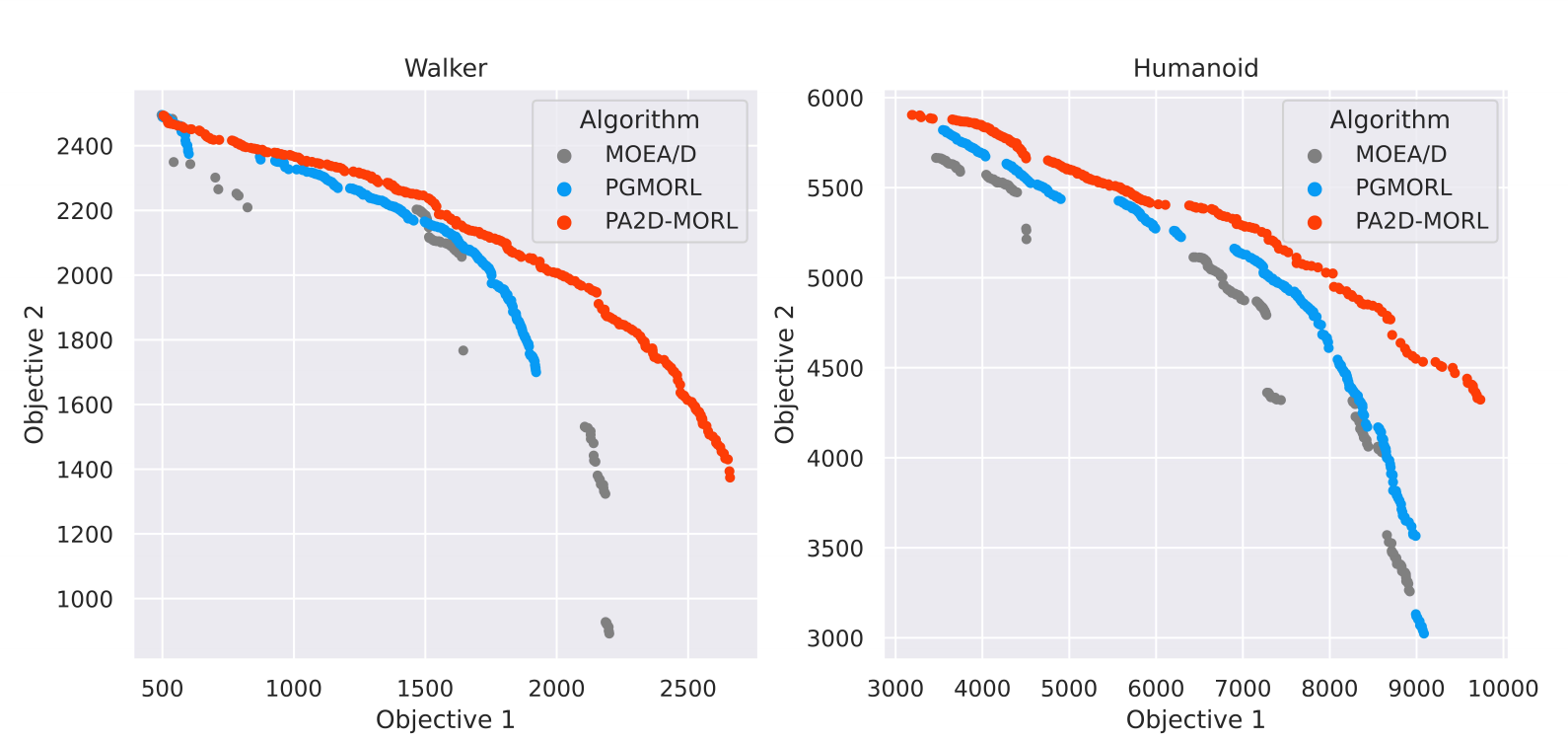} % Reduce the figure size so that it is slightly narrower than the column.
    \caption{Comparison of Pareto frontier approximations. Results of PA2D-MORL, PGMORL, and MOEA/D are shown. PA2D-MORL achieves a higher-quality policy set.}
    \label{points}
  \end{figure}

  \subsection{Simulation Environment}
  
  The proposed method is evaluated in seven multi-objective continuous robot control environments. These environments are based on MuJoCo \cite{todorov2012mujoco}, which is a widely used complex Deep RL benchmark. The state space typically contains the torso position, torso velocity, joint angles, joint angular velocities of robots, and other information from contact sensors. The action space consists of the torques applied at each hinge joint. The original tasks were modified into multi-objective problems with multiple conflicting objectives \cite{xu2020prediction}. For most environments, we use two objectives, forward speed, and energy efficiency. 
  The speed reward is of the form: 
  \begin{equation}
   r_{v} = v + r_{alive},
  \end{equation}
  where $r_{alive}$ is the alive reward, which is added to rewards for all objectives. 
  Energy efficiency is defined as the negative sum of squares of the actions: 
  \begin{equation}
    r_{e} = -\sum_{i} a_{i}^{2} + r_{alive} + C,
   \end{equation} 
  where $C$ is a constant that shifts the reward distribution into positive range. For Hopper-3, there is a third objective, which is the jump height:
  \begin{equation}
   r_{e} = \Delta h + r_{alive} + C,
  \end{equation}
  where $\Delta h $ is the difference between the jump height and the initial height. 
  
  Details about the experimental environment, objectives, and rewards settings are provided in the Appendix.

  \begin{figure}[t]
    \centering
    \includegraphics[width=0.97\columnwidth]{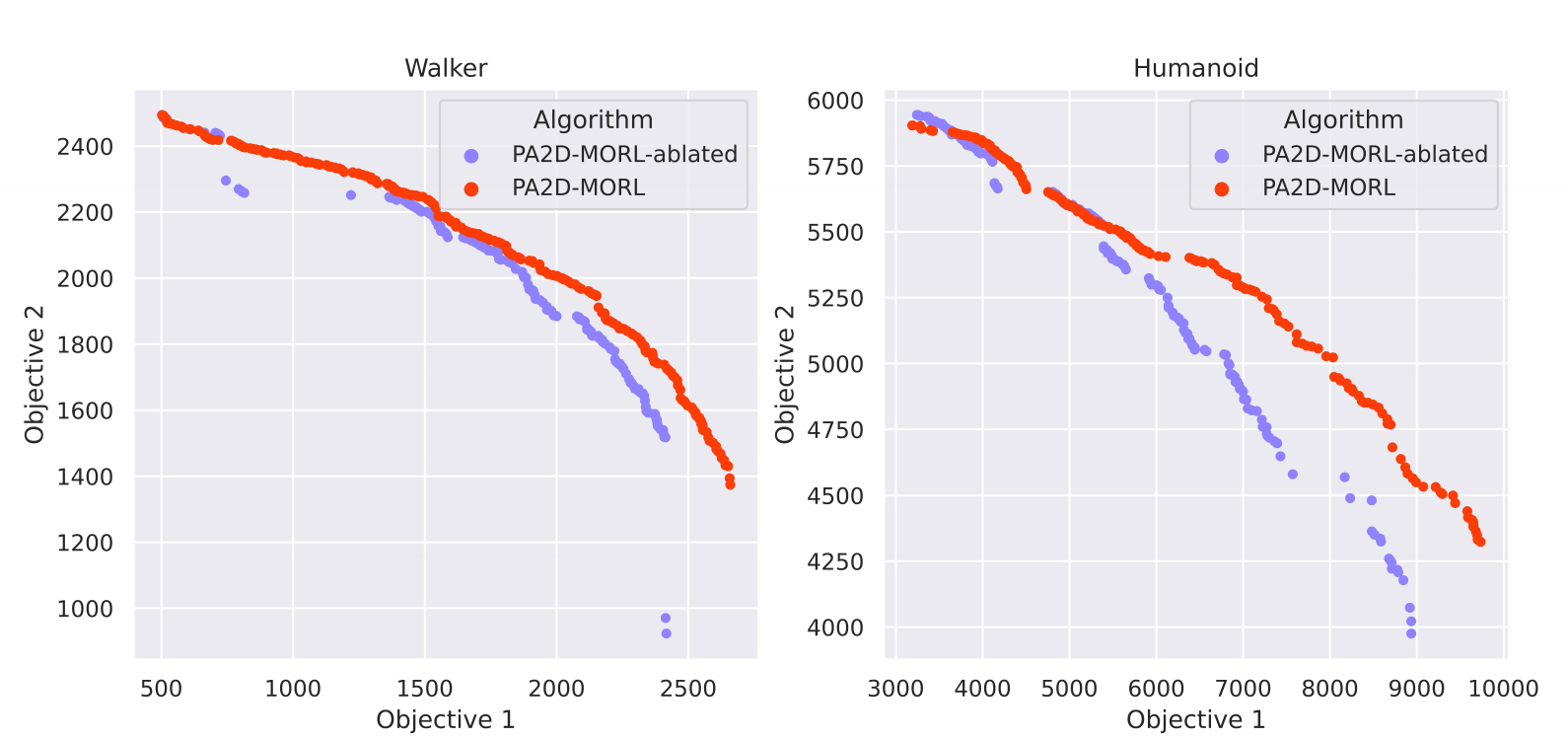} 
    \caption{Comparison of Pareto frontier approximations. Results of PA2D-MORL and PA2D-ablated are shown. PA2D-MORL achieves a denser policy set.}
    \label{points_ablated}
  \end{figure}

  \subsection{Baselines}
  
  We performed ablation experiments without introducing PA-FT during training, which was notated as PA2D-MORL-ablated. In addition, we evaluate three existing multi-policy MORL algorithms: PGMORL, MOEA/D \cite{zhang2007moead}, and PFA \cite{parisi2014policy}. PGMORL is the state-of-the-art MORL algorithm for solving continuous state-action space problems. Meanwhile, existing evaluations have shown that MOEA/D and PFA exhibit good performance in most tasks. Therefore, we include them in our experiments.
  Note that MOEA/D and PFA are a modified version \cite{xu2020prediction} based on the evolutionary framework, as the original algorithms cannot be used directly in the DRL domain. 
  More details about the baseline algorithms are provided in the Appendix.
  
  All baselines, as well as our methods, update 8 policies in parallel within the evolutionary framework, i.e., $p=8$ in Algorithm~\ref{alg1}. The policy parameter updates are performed using the PPO algorithm. Also, all methods contain a warmup phase of $m_w$ iterations, which generates the first generation of policy population. All evaluations and comparisons were based on the same environments and reward settings. Details of the parameters are presented in the Appendix.

  \subsection{Results}
  
  Figure \ref{hv_sp} shows the hypervolume and sparsity curves during training in the Walker environment. 
  Table \ref{allresults} shows the results of the evaluations in all environments, including the average and standard deviation of the HV and SP metrics. All data are based on 6 independent runs. 
  
  In terms of the HV metric, PA2D-MORL has the best results in all environments, suggesting that it generates higher-performance policies and higher-quality Pareto policy set approximations. For the SP metric, PA2D-MORL achieves the best results in most environments except HalfCheetah, in which it has a slight disadvantage compared to PGMORL. This shows that our method can generate a denser policy set. Furthermore, the results of PA2D-MORL are more stable, which has lower standard deviation. 
  
  PGMORL is significantly outperformed by PA2D-MORL in some environments, such as Humanoid and Walker2d. This is possibly because the prediction model tends to get stuck in some local optimums and is unable to guide policies to a higher performance space. The stability of results is also negatively affected by the uncertainty introduced by the prediction model. 
  
  MOEA/D and PFA generate policy sets with high SP metrics, which suggests that the policies generated by these methods are unevenly distributed in certain regions of the objective space. Figure \ref{points} provides the visualized comparison about the distribution of the policy set in the objective space. Note that MOEA/D generates better policies than PGMORL in certain regions of the objective space (e.g., the bottom right corner), but misses many other regions. 
  
  PA2D-ablated shows good performance on HV metrics, with a slightly higher level of sparsity. This demonstrates the importance of introducing the PA-FT approach, which helps to achieve a denser policy set. Figure \ref{points_ablated} shows the comparison between PA2D-MORL and PA2D-ablated in the Walker environment. Note that PA2D-MORL achieves better coverage of the Pareto frontier.

  \begin{table}[t]
    \centering
    \begin{tabular}{c c c c}
      \toprule
      $p$ & metrics & Walker2d & Humanoid \\
      \midrule
        \multirow{2}{*}{$6$} & HV $\uparrow(\times 10^6)$ & $5.652\pm0.193$ & $50.89\pm3.07$ \\
                             & SP $\downarrow(\times 10^4)$& $0.017\pm0.009$ & $0.130\pm0.053$ \\
      \midrule
        \multirow{2}{*}{$8$} & HV $\uparrow(\times 10^6)$& $5.743\pm0.121$ & $51.23\pm2.66$\\
                             & SP $\downarrow(\times 10^4)$& $0.014\pm0.006$ & $0.133\pm0.031$\\  
      \midrule
        \multirow{2}{*}{$10$} & HV $\uparrow(\times 10^6)$& $5.774\pm0.140$ & $52.37\pm2.31$\\
                             & SP $\downarrow(\times 10^4)$& $0.010\pm0.003$ & $0.105\pm0.044$\\  
  
      \bottomrule
    \end{tabular}
  \caption{Comparison of PA2D-MORL with different $p$.}
  \label{parameter p}
  \end{table}

  \begin{table}[t]
    \centering
    \begin{tabular}{c c c c}
      \toprule
      $M_{ft}$ & metrics & Walker2d & Humanoid \\
      \midrule
        \multirow{2}{*}{$\frac{1}{3} M$} & HV $\uparrow(\times 10^6)$& $5.743\pm0.121$ & $51.23\pm2.66$\\
                             & SP $\downarrow(\times 10^4)$& $0.014\pm0.006$ & $0.133\pm0.031$\\ 
      \midrule
        \multirow{2}{*}{$\frac{1}{2} M$} & HV $\uparrow(\times 10^6)$ & $5.761\pm0.317$ & $51.10\pm2.87$ \\
                             & SP $\downarrow(\times 10^4)$& $0.022\pm0.013$ & $0.138\pm0.028$ \\
      \midrule
        \multirow{2}{*}{$\frac{4}{5} M$} & HV $\uparrow(\times 10^6)$ & $5.437\pm0.259$ & $44.86\pm4.83$ \\
                             & SP $\downarrow(\times 10^4)$& $0.131\pm0.090$ & $0.271\pm0.109$ \\
  
      \bottomrule
    \end{tabular}
  \caption{Comparison of PA2D-MORL with different $M_{ft}$.}
  \label{parameter paft}
  \end{table}

  \subsubsection{Parameter Analysis}
  There are two parameters in Algorithm~\ref{alg1} that may affect the results: the number of parallel policies $p$, and the number of generations $M_{ft}$ that determine when the PA-FT is involved. We test different values of these parameters.
  
  We test different values of $p$ in Walker2d and Humanoid environments, and the results are shown in Table \ref{parameter p} . There are slight improvements in the results as $p$ increases. However, this effect is limited, and the performance of the algorithm depends more on the effectiveness of the decomposition and policy selection than on the number of parallel policies. 
  
  $M_{ft}$ determines when we introduce PA-FT into training process. In the experiments above, we use $M_{ft}=\frac{1}{3} M$, meaning that PA-FT is involved at one-third of the training. We also test with other values, which are shown in Table \ref{parameter paft}. The performance of the algorithm decreases significantly only when $M_{ft}$ is increased to $ \frac{4}{5} M$. This suggests that PA-FT should not be involved too late, otherwise it cannot work effectively.  However, $M_{ft}$ is not a sensitive parameter and does not require a lot of tuning.

  \subsubsection{Discussion}
  It is necessary to give some discussion about the theoretical limitations. First, the theoretical analysis of Pareto-stationarity and Pareto ascent direction is founded on convex optimization. For non-convex problems, gradient descent-based optimization may converge to a local optimum, which is a common problem in deep learning. Nevertheless, by stochastic gradient descent, neural networks are usually able to converge to a good enough solution. In addition, through greedy randomized policy selection and multi-generation policy updates, policies are trained with different optimization directions from different parameters, which also helps to avoid getting stuck at the same local optimum. 
  
  Second, the Pareto frontier may be non-convex. Theoretically, a weighted sum-based decomposition cannot reach the non-convex part. Although our method utilizes weighted multi-objective policy gradients, the non-dominated set is established based on Pareto dominance rather than weighted sums, thus policies in non-convex spaces are not eliminated. In addition, the PA-FT approach can also fine-tune policies to the non-convex space. Furthermore, we believe that exploring other ways, such as Tchebycheff approach, is a promising direction for future work.

  \section{Conclusion}
  
  In this paper, we propose PA2D-MORL, a novel multi-policy MORL method that aims to obtain high-quality Pareto policy set approximations in complex DRL domains. 
  An efficient multi-objective decomposition and policy improvement framework is constructed via Pareto ascent directional decomposition and partitioned greedy randomized policy selection. 
  Furthermore, Pareto adaptive fine- tuning is employed to improve the density and spread of Pareto approximations. Evaluations in various environments show that PA2D-MORL outperforms existing methods in terms of both the quality of the Pareto approximation and the stability of the results, demonstrating the effectiveness of the method. 
  
  We expect that most policy-based DRL methods can be integrated with our approach and thus applied to more domain-specific problems. In addition, it is promising to combine this approach with safe reinforcement learning to tackle some constrained decision-making problems.

  \section{Acknowledgments}
This work was supported in part by the National Natural Science Foundation of China under Grants 62373375, U2341216 and 62022094, in part by the Zhejiang Lab under Grant 2021NB0AB01.

\bibliography{aaai24}

\end{document}